%% file: ms.tex
\documentclass{IEEEtran}
\usepackage{cite}
\usepackage{amsmath,amssymb,amsfonts}
\usepackage{algorithmic}

\usepackage{graphicx}
\usepackage[export]{adjustbox}

\usepackage{textcomp}
\usepackage[table,xcdraw]{xcolor}
\definecolor{ToDoColor}{HTML}{00b503}
\definecolor{green}{HTML}{008947}
\definecolor{orange}{HTML}{C56E00}

\usepackage{cancel}
\usepackage{csquotes}
\usepackage{booktabs}
\usepackage{makecell} 
\usepackage{multirow}

\usepackage[list=off]{subcaption}
\usepackage[hyphens]{url}
\hypersetup{
  colorlinks   = true,             
  breaklinks   = true,             
  urlcolor     = {blue!80!black},  
  citecolor    = {green!60!black},  
}

\graphicspath{{./images/}}
\setkeys{Gin}{width=\linewidth} 
\newcommand{\fincludegraphics}[2][]{\frame{\includegraphics[#1]{#2}}}

\usepackage[T1]{fontenc}
\usepackage{pifont}
\newcommand{\cmark}{\textcolor{green}{\ding{52}}}
\renewcommand{\o}[1]{\textcolor{orange}{#1}}

\newlength{\myfigheight} 
\begin{document}
\renewcommand*{\figureautorefname}{Fig.}

\title{Field typing for improved recognition on heterogeneous handwritten forms}

\author{
	Ciprian Tomoiag\u{a}$^{1*}$, Paul Feng$^{1*}$, Mathieu Salzmann$^2$, Patrick Jayet$^1$  \vspace{0.1cm}\\
	$^1$ AXA REV Lausanne\\
	$^2$ CVLab, EPFL, Switzerland  \vspace{0.1cm}\\
	\texttt{\small\{ciprian.tomoiaga, paul.feng, patrick.jayet\}@axa.com, mathieu.salzmann@epfl.ch}
	\\
	{\small * equal contributions}
}

\IEEEpubid{2379--2140/19/\$31.00~\copyright~2019 IEEE}
\markboth{2019 International Conference on Document Analysis and Recognition}{}

\maketitle

\begin{abstract}
Offline handwriting recognition has undergone continuous progress over the past decades. However, existing methods are typically benchmarked on free-form text datasets that are biased towards good-quality images and handwriting styles, and homogeneous content. In this paper, we show that state-of-the-art algorithms,  employing long short-term memory (LSTM) layers, do not readily generalize to real-world structured documents, such as forms, due to their highly heterogeneous and out-of-vocabulary content, and to the inherent ambiguities of this content. To address this, we propose to leverage the content type within an LSTM-based architecture.
Furthermore, we introduce a procedure to generate synthetic data to train this architecture without requiring expensive manual annotations. We demonstrate the effectiveness of our approach at transcribing text on a challenging, real-world dataset of European Accident Statements.
\end{abstract}


\section{Introduction}
As they adapt their processes to the digital world, many companies have a need for automatically extracting information from digitized documents. For printed documents, advances in OCR software have almost solved the problem. Similarly, fully handwritten ones can now be processed with high recognition rates~\cite{dutta2018_hwr}, even above human performance for some tasks~\cite{oliveira_superhuman}, thanks to the strong focus of the research community. However, as noted in~\cite{eskenazi2017comprehensive}, the data used for research purposes is strongly biased towards newspapers and journals, and in both cases mostly written in English. As a consequence, existing techniques are ill-suited to handle structured forms, such as accident statements, where printed text prompts handwritten input. The reasons for this are threefold, as discussed below.

\begin{figure}
    \setlength{\myfigheight}{0.42\textheight}

    \begin{minipage}[c][\myfigheight]{.5\linewidth}
		\includegraphics{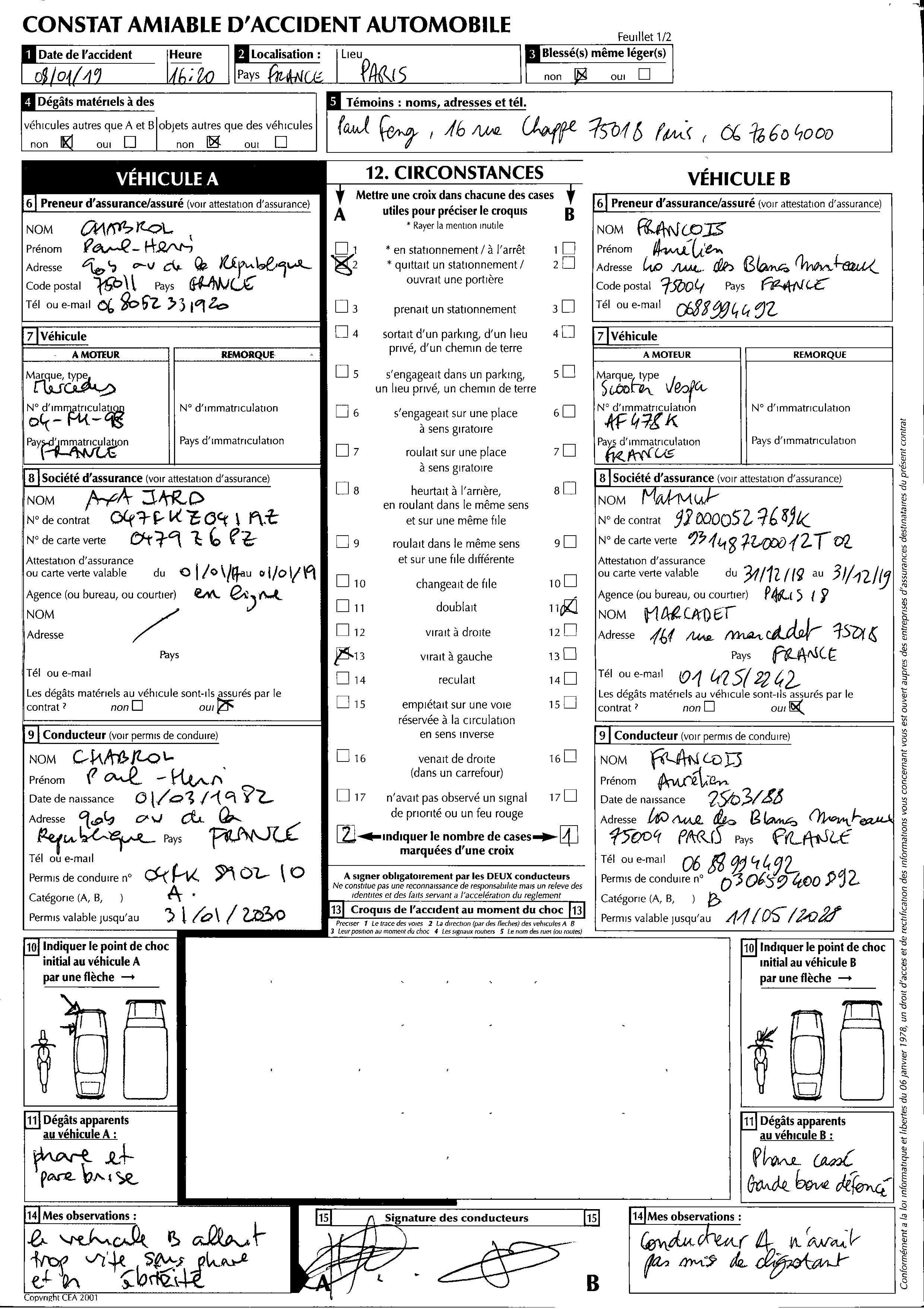}
		\subcaption{Atypical structure of the form.}
		\label{fig:constat_full}
		\vfill

		\fincludegraphics{real/overlap2}%
		\vspace*{.5em}
		\fincludegraphics{real/overlap}
		\subcaption{Overlaps due to limited space.}
		\label{fig:constat_extracts:overlaps}
    \end{minipage}
    \hfill
	\begin{minipage}[c][\myfigheight]{.45\linewidth}
        \fincludegraphics{real/1}%
        \vspace{.5em}
        \fincludegraphics{real/outside}%
        \vspace{.5em}
		\fincludegraphics{real/abbrev}
		\subcaption{Crammed handwriting and unusual abbreviations.
		}
		\label{fig:constat_extracts:crammed}
		\vfill

        \vspace{1em}
	    \fincludegraphics{real/947_modif}%
	    \vspace{.5em}
		\fincludegraphics{real/881}
		\subcaption{Degradation and noise.}
		\label{fig:constat_extracts:noisy}
		\vfill

		\begin{center}
        \includegraphics[height=.72cm,keepaspectratio=true]{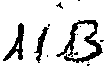}

		\includegraphics[height=.72cm,keepaspectratio=true]{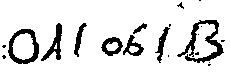}
		\end{center}
		\subcaption{Knowing the content type and context is crucial to disambiguate some glyphs.}
		\label{fig:constat_extracts:context}
	\end{minipage}

	\caption[nothing]{European Accident Statement and its challenges.}
	\label{fig:EAS_example}
\end{figure}

First, as shown in~\autoref{fig:constat_full}, such documents comprise highly heterogeneous content, where printed and handwritten text are mixed, and out-of-vocabulary (OOV) instances, such as names and surnames, phone numbers and dates, dominate. This contrasts with \emph{free-form} text databases constructed specifically for handwriting recognition (HWR), such as IAM~\cite{iam} or RIMES~\cite{rimes}, which, although realistic, contain only correct sentences,
thus allowing HWR systems to benefit
from language models, whether explicitly~\cite{explicit_LM_2015} or implicitly~\cite{implicit_LM_2018}.

Second, forms such as that of~\autoref{fig:constat_full} present an atypical structure that is not based on text lines, paragraphs or columns. In particular, the limited input space often results in the handwritten text overlapping with the printed one or the form's separators, as depicted by~\autoref{fig:constat_extracts:overlaps}, and forces the writer to cram their handwriting or use non-standard abbreviations, as illustrated in~\autoref{fig:constat_extracts:crammed}. Furthermore, a writer often does not use a consistent style throughout the form, mixing ligature handwriting with hand print, which further complicates the separation of user text from form one.

Third, in practice, the documents to be processed are often of much poorer quality than those in academic databases. This is due to the lack of control of companies on the digitization process, which results in them receiving poor-quality scans or faxes of \emph{carbon-copy} forms, already thresholded with less than ideal parameters, as shown in~\autoref{fig:constat_extracts:noisy}.

Altogether, the three factors mentioned above lead to dramatically underwhelming HWR performances. Specifically, training a state-of-the-art CNN-RNN architecture~\cite{CRNN} on the French RIMES dataset and applying it on French accident statements, such as that of~\autoref{fig:constat_full}, results in mostly wrong transcriptions, with an average character error rate (CER) of 71\%. While this suggests a clear domain shift between RIMES and accident forms, we observed that these results are not only due to this shift; in some cases, the transcriptions are optically accurate, as the glyphs can have a double interpretation even for humans. For example, one would probably read the top portion of \autoref{fig:constat_extracts:context} as the \emph{letters} ``\texttt{AIB}'', whereas the glyphs were copied from the \emph{date} ``\texttt{01/06/13}'' in the bottom portion of the figure. However, knowing the type of this data, a human would never make this mistake.

\subsection{Contributions}
\IEEEpubidadjcol 

In this paper, we introduce an approach to performing HWR on structured and heterogeneous forms. Motivated by our previous observation, we propose to improve the performance of state-of-the-art HWR systems by explicitly reasoning about the type of the observed content. Specifically, we input an additional feature indicating this type into our CNN-RNN model. This allows the recurrent layers to learn multiple distributions over the input sequence and switch between them as necessary, thus helping the network choose the correct representation for an ambiguous glyph. Note that this contrasts with typical CNN-RNN architectures, including those used for image-captioning, which \emph{implicitly} encode image priors via the convolutional features.

Furthermore, to tackle the fact that training on public databases generalizes poorly to structured forms, we introduce an automatic approach to generating synthetic training data. To this end, we leverage empty form templates that we complete with handwritten text so as to simulate users' inputs and thus create documents with highly heterogeneous content. In this process, we keep track of the data type in each field of the form, so as to further mimic ambiguous situations and force the network to exploit the input type. Finally, we apply randomized transformations and degradations to the image to reflect the high degree of appearance diversity that can be observed in practice. Altogether, this allows us to create thousands of images without any need for manual annotation.

We demonstrate the benefits of our approach on a dataset of 86  real European Accident Statements (EAS), from which we extracted approximately 4200 handwritten fields for evaluation purposes. As evidenced by our empirical results, the use of our synthetic data together with accounting for the fields' type allows us to reduce the CER from 71\% for the original model trained on RIMES to 23\% with our final model. Our code and synthetic dataset is available at: \url{https://github.com/cipri-tom/type-aware-crnn}

\input{related}

\begin{figure}[t]
    \centering
    \includegraphics{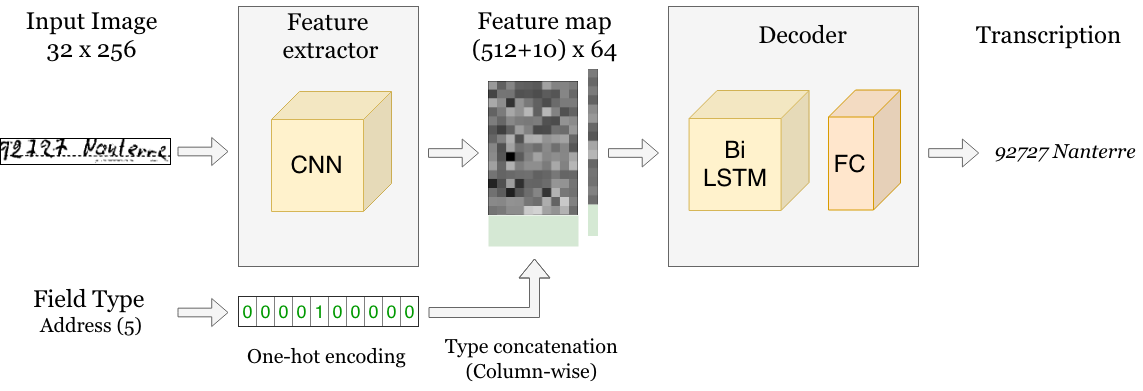}
    \caption{Overview of our type-aware transcription architecture.}
    \label{fig:crnn_overview}
\end{figure}

\section{Methodology}
In this section, we introduce our framework for HWR from structured forms. We first present our type-aware recognition strategy and then our approach to creating synthetic data. Finally, we summarize the entire pipeline, including detection of handwritten text and \emph{automatic} type determination for each detected field.

\subsection{Type-aware HWR Architecture}
\label{sec:architecture}
Following the recent advances in HWR, we base our approach on the CNN-RNN architecture of~\cite{CRNN}. However, to account for the fact that some glyphs are ambiguous unless we know what type of content they represent, we further incorporate an additional input encoding the content type. As shown in \autoref{fig:crnn_overview}, the resulting architecture accepts as input an image of 32 pixels of height and variable width together with a one-hot encoding of the content type, and produces a text sequence as output.

The image is passed through a VGG-inspired feature extractor with 7 convolutional layers of kernel size \(3 \times 3\) and with ReLU activation functions. These are interspersed with 3 BatchNorm layers and 4 MaxPool layers. Note that only 2 of the MaxPool layers downscale the image horizontally (i.e., have a \(2 \times 2\) receptive field), while the other 2 only downscale it vertically (i.e.,  have a \(2 \times 1\) receptive field). This allows us to retain information about thin characters, such as \(\{l, i, 1, / \}\).

The content type is then concatenated with the convolutional features of each column, and the resulting features are fed as a sequence to the recurrent part. This recurrent part comprises 2 layers of \mbox{bi-LSTM} cells, each with hidden state of size 256. Finally, the output at each column step is mapped to a probability distribution over the 70 symbols of the alphabet via a fully-connected layer.

\subsection{Synthetic Training Data}
\label{sec:synthetic_data}

Getting sufficiently many annotated samples to train a deep network is typically very expensive. In the context of personal forms, such as accident statements, this problem is further increased by the high sensitivity of the content. In principle, one could of course use general-purpose external data, such as images from the RIMES or the IIIT-HWS databases. However, as discussed above, these datasets do not reflect our final goal, particularly in terms of content and  diversity, and thus a model trained using this type of data will generalize poorly to structured forms.

To overcome this, we therefore introduce a synthetic data generator that renders strings with a high degree of variability based on different sources. These include generators of numbers, dates, times, addresses, names and correct license plates. Since each source is linked to a specific content type, our data generator directly provides us with all the necessary information to train our network. Note that this would not be the case if using standard datasets, for which content types are not available.
To avoid overfitting on any content type, we need the distribution of the generated data to match that of real forms.
To this end, we therefore estimate the proportion of each field type based on a template form and perform weighted sampling accordingly.

\begin{figure}[t]
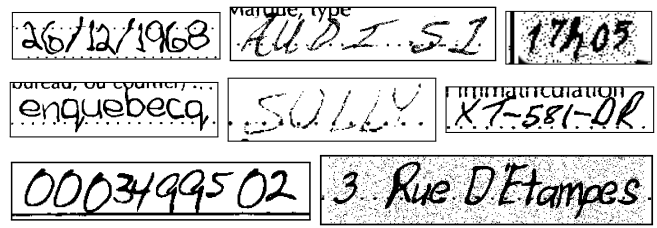

    \centering
    \setkeys{Gin}{width=.31\linewidth}
    \adjustboxset{frame,valign=c}
    \adjincludegraphics{synth/date_elastic}
    \adjincludegraphics[width=.4\linewidth]{synth/car_form}
    \adjincludegraphics[width=.22\linewidth]{synth/time}%
    \vspace{.5em}

    \adjincludegraphics{synth/name_form}
    \adjincludegraphics{synth/vocab_degraded2}
    \adjincludegraphics{synth/plate_modif}%
    \vspace{.5em}

    \adjincludegraphics[width=.45\linewidth]{synth/number}
    \adjincludegraphics[width=.5\linewidth]{synth/address}
    \caption{Examples of synthetic data.}
    \label{fig:synthetic_data}
\end{figure}

To cover a wide enough appearance variation, we use a collection of 800 fonts that resemble handwriting, with random amounts of kerning and vertical displacement between the letters. We then further apply different deformations to the resulting images: First, we use affine transformations to simulate rotation and slanting of the handwriting, while generating different image scales. Since the documents are scanned flat, there is no need to include projective transformations. Second, we simulate the uneven forces in the human hand muscles by implementing elastic distortions, as described in~\cite{elastic_distortions} for MNIST digit recognition. We sample the elasticity coefficient \(\sigma \) from a normal distribution \(\mathcal{N}(8, 2)\) and set the intensity \(\alpha\) to the height of the text. Third, we degrade the images using erosion, dilation, gradient and closing operations with random structure elements, which produces realistic noise. Finally and most importantly, to simulate the overlapping of input data over form elements, we paste the images into the designated input spaces of a template form and crop out a slightly larger region around them. Some examples of generated images are shown in \autoref{fig:synthetic_data}.

\begin{table}[t]
\centering
\caption{Distribution of content types in real data.}
\label{tab:type_distribution}
\begin{tabular}{@{}lrr@{}}
    \toprule
    \textbf{Content type} & \textbf{Sample size} & \textbf{Fraction} (\%) \\ \midrule
    Free Text & 1181 & 28.49 \\
    Name  & 594  & 14.33 \\
    Phone Numbers & 241  & 5.81 \\
    Date  & 435  & 10.49 \\
    Time & 75   & 1.81 \\
    Address  & 805  & 19.42 \\
    License Plate & 141  & 3.40  \\
    Numbers  & 335  & 8.08 \\
    Car Model & 129  & 3.11 \\
    Insurance Name & 210  & 5.07 \\
    \midrule[0.025em]
    \textbf{TOTAL} & 4146 & 100\\
    \bottomrule
\end{tabular}
\end{table}

\subsection{Overall Pipeline}
\label{sec:method:system}

At test time, given a new, real document, we need to extract the individual fields and their type automatically to then feed each of them to our HWR architecture. To achieve this, we develop a pipeline consisting of three steps. Note that this pipeline, as our data generation procedure discussed above, relies on a template form, that is, a blank form, whose input zones are tagged with their respective content type. Computing this template, however, is a one-time process for a given form.

In the first step of our pipeline, we approximately align the template form to the input document. This is similar to what was done in~\cite{skew_correction_2016} via the probabilistic Hough transform and in~\cite{aldavert_dynamic_content} via the Lukas-Kanade registration algorithm. Here, we rely on detecting squares in the two images. The positions of the squares are then used as feature points and the alignment transformation is estimated using an Iterative Closest Point algorithm~\cite{besl1992_ICP}. Note that this strategy can also serve to discard non-conforming documents.

As a second step, we detect handwritten text using the Faster \mbox{R-CNN} architecture~\cite{faster_rcnn}. This allows us to find tight bounding boxes around text even if it does not conform to the form boundaries. We trained this network using the synthetic data produced with the generator discussed above. In practice, this technique successfully extracts 92\% of the handwritten text regions in real documents.

Finally, as a third step, we assign a type to each detected handwritten text region based on its maximum overlap with a form region. We then crop the detected region from the image and, together with its identified type, pass it to the network described in Section~\ref{sec:architecture} for transcription.

\section{Experiments}
\label{sec:experiments}

We now evaluate our framework on a challenging dataset of EAS images.
To the best of our knowledge, no other works have explicitly tackled HWR for such structured and heterogeneous documents. Therefore, we make use of the state-of-the-art CRNN architecture of~\cite{CRNN}, which serves as backbone to our approach, as a baseline. Below, we first present our experimental setup, then discuss our results, and finally analyze the behavior of our network.

\subsection{Experimental setup}
As the text images extracted from an input form may be of arbitrary size, we scale them to 32 pixel of height. Images whose post-scaling width is shorter than 256 pixels are padded to the right with copies of themselves, while keeping track of the original width. This allows us to arrange them in mini-batches while keeping a constant distribution of black and white pixels across batches.

For training, we use 2 million synthetic examples in total, organized in mini-batches of 512 examples. We employ the alignment-free CTC loss function of~\cite{CTC}, and we optimize the network parameters using Adam~\cite{adam} with an exponential learning rate \(\alpha = 0.001\), decreasing every 5000 iterations.

We check the convergence and generalization of any model using a validation set. For models trained on RIMES, it consists of the official RIMES test set. For models trained on our synthetic data, the validation set is taken as a separate synthetic dataset rendered using 100 fonts that were not seen during training.

For evaluation, we manually extracted and transcribed $\sim$\,4200 handwritten fields from French EAS forms, which allows us to evaluate text recognition in isolation of previous parts of the pipeline. These fields were then categorized into 10 types, the distribution of which is detailed in~\autoref{tab:type_distribution}. For our comparisons to be fair, all models are evaluated on this dataset using the average Character Error Rate (CER) and  Field Error Rate (FER) defined as
\[
\begin{aligned}
	\operatorname{CER} &= \frac{100}{N} \sum_i \frac{\operatorname{EditDist}(\mathit{prediction}_i, \mathit{groundtruth}_i)}{\operatorname{NumChars}(\mathit{groundtruth}_i)},\\
	\operatorname{FER} &= \frac{100}{N} \sum_i (1 - \delta (\mathit{prediction}_i, \mathit{groundtruth}_i)),
\end{aligned}
\]
where \(i\) indexes over the samples from our evaluation dataset and \(\delta\) outputs 1 if its arguments are the same, letter for letter, or 0 if they differ. While FER has the same definition as the commonly used Word Error Rate, note that we compute it over an entire input \emph{field}, which can contain multiple words. In our results below, we also report CER-ASCII and FER-ASCII metrics, which are essentially the same as those above, except that they do not penalize some acceptable confusions, such as {\it e} for {\it \'e}.

\subsection{Results}

\begin{table}[t]
\centering
\caption{Overall results.}
\label{tab:final_results}
\renewcommand{\arraystretch}{1.2}
\begin{tabular}{@{}lcccc@{}}
    \toprule
    \textbf{Model} trained on: & \textbf{CER} & \textbf{\makecell{CER\\ASCII}} & \textbf{FER} & \textbf{\makecell{FER\\ASCII}} \\\midrule
    RIMES (untyped) & 70.7 & 70.3 & 97.7 & 97.2 \\
    Synth (untyped) & 51.8 & 50.2 & 90.7 & 89.2 \\
\midrule[0.025em]
    Synth + Type & 46.2 & 44.9 & 88.7 & 87.3 \\
    Synth + Type + Augment & 22.8 & 21.8 & 68.9 & 66.4\\\bottomrule
\end{tabular}
\end{table}

\begin{figure}[t]
\centering
\resizebox{\linewidth}{!}{%
\begin{tabular}{@{}c@{\hspace{1ex}}cll@{}}%
    \textbf{ID} & \textbf{Image} & \textbf{Model}* & \textbf{Transcription}\\\midrule

    \multirow{5}{*}{(a)\phantomsubcaption\label{fig:results:a}}
    & \multirow{5}{*}{\adjincludegraphics[width=4cm]{real/2326}}
	& \textbf{R} & Jergeatzne \\
	& & \textbf{S} & Beugeot 3061 \\
	& & \textbf{ST} & Peugeot 3061 \\
	& & \textbf{STA} & Peugeot 306 \cmark \\\cmidrule[0.025em]{3-4}

	\multirow{4}{*}{(b)\phantomsubcaption\label{fig:results:b}}
	& \multirow{4}{*}{\adjincludegraphics[width=4cm]{real/290}}
	& \textbf{R} & IlAueislesilasilis \\
	& & \textbf{S} & Ay.12u.elslesilasitiis- \\
	& & \textbf{ST} & 1u, Bue'uesiRasizis \\
	& & \textbf{STA} & 10  Rue Des Rosiers \cmark \\\cmidrule[0.025em]{3-4}

	\multirow{4}{*}{(c)\phantomsubcaption\label{fig:results:c}}
	& \multirow{4}{*}{\adjincludegraphics[width=4cm]{real/context_date}}
	& \textbf{R} & 01T06IT \\
	& & \textbf{S} & 04/06/ \\
	& & \textbf{ST} & 01/06/83 \\
	& & \textbf{STA} & 01/06/13 \cmark \\\cmidrule[0.025em]{3-4}

	\multirow{5}{*}{(d)\phantomsubcaption\label{fig:results:d}}
	& \multirow{5}{*}{\adjincludegraphics[width=4cm]{real/overlap2}}
	& \textbf{R} & iteAuterejeral \\
	& & \textbf{S} & VoAfmBerojerac \\
	& & \textbf{ST} & MqoAm Bergierac \\
	& & \textbf{STA} & A\o{x}A Bergera\o{l} \\
	& & \textbf{GT} & AXA Bergerac \\\cmidrule[0.025em]{3-4}

	\multirow{5}{*}{(e)\phantomsubcaption\label{fig:results:e}}
	& \multirow{5}{*}{\adjincludegraphics[width=4cm]{real/881}}
	& \textbf{R} & leletetetetetestlete \\
	& & \textbf{S} & 6800001980 \\
	& & \textbf{ST} & 22.3.9F9019391 \\
	& & \textbf{STA} & 26/1\o{4}/19\o{8}1 \\
	& & \textbf{GT} & 26/11/1991 \\\cmidrule[0.025em]{3-4}

	\multirow{5}{*}{(f)\phantomsubcaption\label{fig:results:f}}
	& \multirow{5}{*}{\adjincludegraphics[width=4cm]{real/408_modif}}
	& \textbf{R} & Co742666040 \\
	& & \textbf{S} & Culnle2n.6869.g0 \\
	& & \textbf{ST} & 00.9.61.91.68.63.80 \\
	& & \textbf{STA} & 0\o{7.}4\o{7.}68\o{.}6\o{8.8}0 \\
	& & \textbf{GT} & 0949686960 \\\cmidrule[0.025em]{3-4}

\end{tabular}%
}%
    \caption{Examples of difficult images and their transcriptions with different models {\footnotesize(R\,=\,RIMES, S\,=\,Synthetic, ST\,=\,Synthetic\,+\,Type, STA\,=\,Synthetic\,+\,Type\,+\,Augment, GT\,=\,Ground Truth)}. The synthetic data (S) already helps recognize some easier OOV instances, as in \subref{fig:results:a} and \subref{fig:results:c}. Exploiting types (ST) further improves cases such as \subref{fig:results:c} and \subref{fig:results:f}, where the model then predicts complete date or digits. Finally, data augmentation (STA) overcomes the problems of irrelevant content in most of the images. Note, however, that, as shown in \subref{fig:results:f}, the typed architecture tends to be biased towards grouped digits for phone numbers, which is a peculiarity of the French language.
    }
    \label{fig:results}
\end{figure}

\begin{figure}[t]
    \centering
    \includegraphics{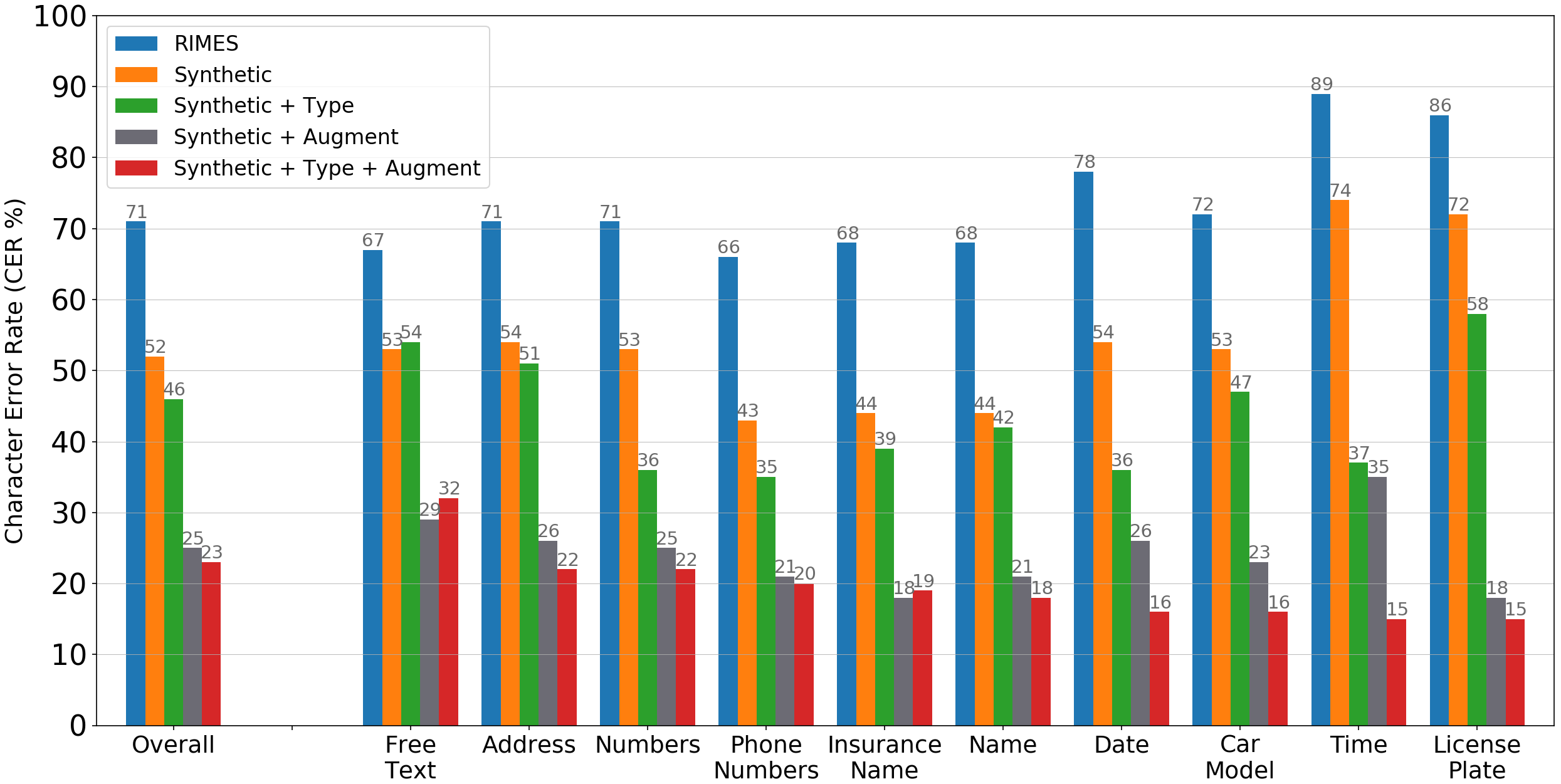}
    \caption{Performance of different models, overall and broken down by content type.}
    \label{fig:comparison_cer}
\end{figure}

The overall results of all evaluated models, including different versions of our approach, are provided in~\autoref{tab:final_results}.
Furthermore, in~\autoref{fig:comparison_cer}, we break down the results by field type, which helps understand the influence of different factors.

Our baseline consists of a vanilla CRNN architecture trained on the RIMES dataset so as to match the language of our EAS documents. While it attains 8\% CER on the RIMES test set, its disappointing performance of 71\% CER on our real data confirms the huge shift in content and style between academic databases and real-world structured forms. Using synthetic training data to address the difference in content decreases the CER by 19pp on average, with a great recognition improvement for number types. However, the resulting model still struggles with OOV instances, such as times and license plates. By contrast, our type-aware architecture further lowers down the error rate by 6pp overall. Importantly, while types such as free text are understandably unaffected by the use of the type as input, better-defined types such as times and license plates undergo a dramatic error drop of more than 37pp for time and 14pp for license plate.

Visual inspection of the model at this stage suggests that the remaining errors are typically due to poor image quality, as shown in~\autoref{fig:results} (a, b, e). This, however, can be largely addressed by our data augmentation strategy described in Section~\ref{sec:synthetic_data}. Doing so allows us to halve the CER, thus reaching 23\% overall, with fields such as license plate, which are crucial to identify an insured driver, reaching a CER as low as 15\% (or 11\% CER-ASCII). For a typical length of 7 characters, this means that on average the system mistakes less than one character per instance. Note also that, with the exception of free text, most fields have similar recognition rates. Importantly, free text typically is the least significant content type for any subsequent processing task; discarding its contribution to the error rate translates to our approach achieving 19\% CER (or 18\% CER-ASCII).

From~\autoref{fig:comparison_cer}, where we study the the impact of the different components of our approach, one can see that our data augmentation schemes significantly contribute to the success of our approach. This is due to the fact that one of the data augmentation techniques consists of rendering text on the template form and cropping a region with its context, which leaks information about the content type. As such SA also has access to knowledge about the type. Nevertheless, explicit use of the field type (STA) still further improves the results.

\subsection{Network Analysis}

We now analyze the behavior of our architecture, with a particular focus on understanding the effect of using the content type as input. To this end, we observe the influence of using a single type as input for all test samples on the output of the architecture.

\begin{figure}[t]
    \centering
    \begin{subfigure}{\linewidth}
        \includegraphics{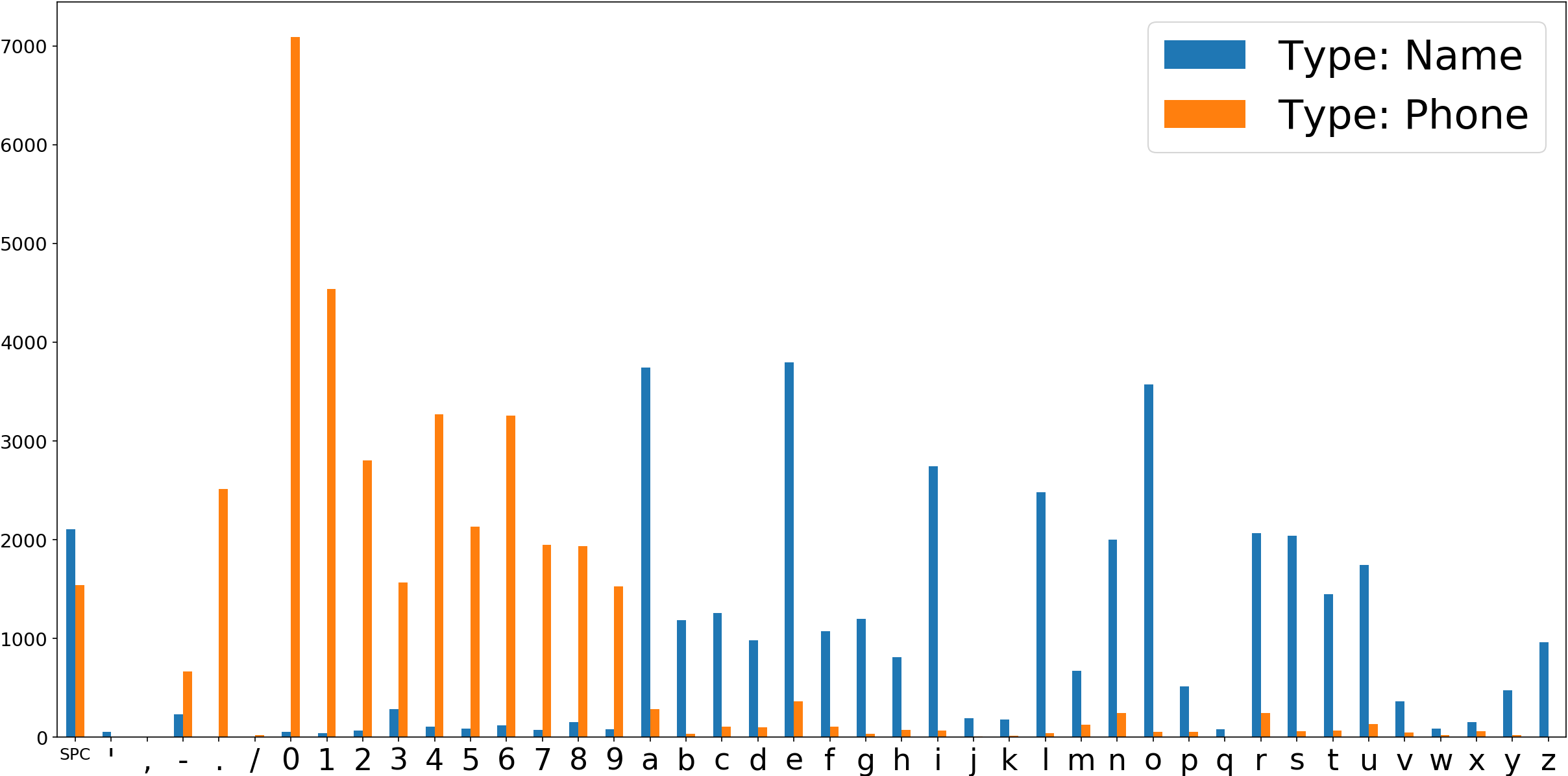}
        \caption{Distribution of predicted symbols when forcing the same type for all examples of the test set. Using \textsl{Phone number} clearly yields more digits, whereas using \textsl{Name} makes the network focus on letters.}
        \label{fig:logits:single_type}
    \vspace{0.2cm}
    \end{subfigure}
    \begin{subfigure}{\linewidth}
        \includegraphics{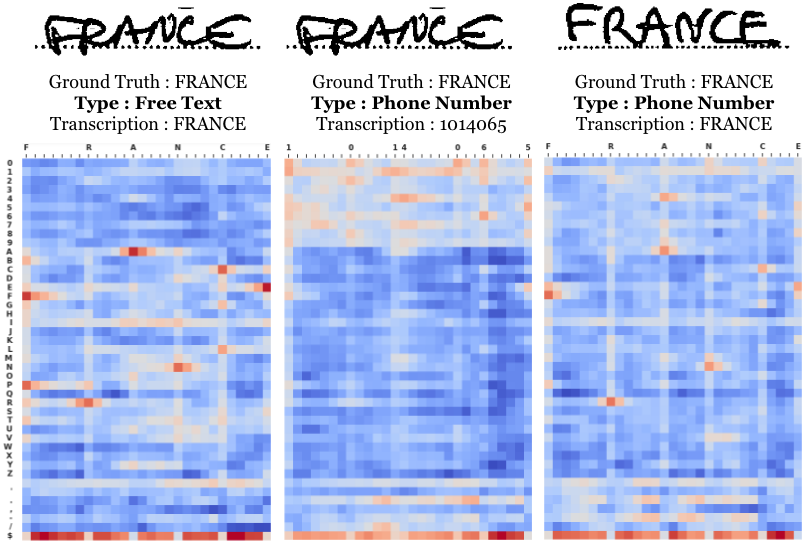}
        \caption{Logits variation for "France" images for different input types but using the same model. The $y$-axis shows the reduced alphabet considered (ordered as: digits + uppercase latin alphabet + special characters + blank). The $x$-axis shows the transcription corresponding to each input (i.e., column) in the sequence. Intensity (min=blue, max=red) reflects the likelihood of a character to be predicted.}
        \label{fig:logits:france}
    \end{subfigure}
    \caption{Influence of the type on the network's outputs.}
\end{figure}

In particular, in \autoref{fig:logits:single_type}, we plot the distribution of predicted symbols over the entire test set when the input type is forced to Name (blue) or to Phone number (orange). While, in the former case, the dominant symbols are clearly the letters, in the latter one, digits are predicted much more frequently. This confirms that the network has indeed learnt to leverage the type information for transcription.

This can be further evidenced by observing the network's logits for individual examples, as illustrated in \autoref{fig:logits:france}, where the same image input in conjunction with different types yields completely different probability distributions, hence transcriptions. Indeed, for the same image, the focus is put on letters when the type is free text (left), as opposed to digits when the type is phone number (middle).
Interestingly, when inputting a different image depicting the same word with the type phone number (\autoref{fig:logits:france} right), the transcription is correct. This evidences that the network relies on the type only when necessary, that is, to disambiguate complex images for which the model is not confident.

\section{Conclusion}
Motivated by the observation that state-of-the-art HWR methods generalize poorly to form-like documents, we have introduced an approach to transcribe structured forms with highly heterogeneous content. In particular, we have proposed to leverage the content type to overcome inherent ambiguities and have introduced a data generation strategy to avoid the expensive manual annotation process. Our results evidence the benefits of the different components of our approach, allowing us to reduce the CER from 71\% to only 23\%. Importantly, adding field type information has enabled us to train a \emph{single} robust and flexible model effective for all types, rather than training a separate model for each type, which is computationally expensive and inefficient. Although our model is trained for pre-defined types, our synthetic data generation procedure is generic, and so re-training for a new field type requires little effort. In the future, we intend to study the use of domain adaptation strategies to further reduce the gap between our synthetic data and real documents.

\bibliographystyle{IEEEtran}
\bibliography{refs}

\end{document}

%% file: related.tex
\section{Related work}

\subsection{Text recognition}
Automatic text recognition has been a longstanding goal of the document processing community, as discussed thoroughly in the survey of~\cite{eskenazi2017comprehensive}. As for many image-analysis problems, Convolutional Neural Networks~\cite{leCun_CNN} have proven highly beneficial for text recognition, starting with the simple task of classifying individual handwritten digits. These results were extended to more general characters~\cite{jaderberg2014_unconstrained}, as well as to sequence-based inputs, such as text, speech and music scores, by exploiting Markov models~\cite{plotz2009markov}. In turn, these sequential models were also replaced with artificial neural networks, in particular recurrent ones, such as the Long Short Term Memory (LSTM) of~\cite{LSTM_original}, which have proven more effective at handling long-term dependencies~\cite{graves_LSTM}. In particular, coupled with the Connectionist Temporal Classification loss, LSTMs constitute a highly-successful, alignment-free recognition method~\cite{CTC}.
Further improvements were made by using dropout within the recurrent network as a mean of regularization~\cite{MDLSTM_dropout}.

Ultimately, the current state-of-the-art architectures for both text-in-the-wild~\cite{CRNN} and handwriting~\cite{rawls2017_CNN,puigcerver_MDLSTM_vs_CNN,dutta2018_hwr} recognition combine CNN and RNN layers so as to model individual characters and sequences thereof in a single framework that can be trained in an end-to-end fashion.
This is typically achieved by using VGG-derived convolutional feature extractors followed by bi-LSTM~\cite{LSTM_original} layers. In addition, these methods usually exploit synthetic data and data augmentation to improve robustness, as well as an external language model to correct the character recognition errors. 
While most methods focus on a single language, the multilingual setting was addressed in~\cite{bluche2017gated} via a new gated convolutional feature extractor.
Note that, while convolutional extractors are the most common ones, fully-connected layers can also be employed, as demonstrated in~\cite{rawls2017_direct}. In any event, the effectiveness of these techniques has been demonstrated on academic databases only, and these databases consist of standard text lines and paragraphs. In other words, as discussed above, state-of-the-art models trained on such academic databases are ill-suited to handle structured forms with heterogeneous content. Here, we introduce an approach to handling such forms, which relies on leveraging the text type to overcome ambiguities and on generating synthetic data to avoid manual annotation.

\subsection{Synthetic data and preprocesing techniques}

Creating a synthetic dataset to compensate for the lack of annotated handwriting data has already been studied in the past. This, for instance, was the case in~\cite{synthetic_data} to generate synthetic natural scene text images and in~\cite{krishnan2016generating} to render 90k English words using a collection of handwritten fonts. In~\cite{generation_nunamaker2016tesseract}, instead of fonts, images of letters were concatenated so as to create a database used to train an OCR engine. None of these methods, however, apply to our scenario, because they generate either printed text only, without ligatures, or words from the English vocabulary, thus being ill-suited to handle OOV instances. Furthermore, they do not consider structured documents as we do here.


When working with such documents, one needs to extract the regions containing handwritten text. This problem, however, is not the main focus of our work, and we refer the reader to~\cite{eskenazi2017comprehensive} for a comprehensive survey of the literature until 2015. More recent methods are largely divided into two classes. On one hand, there are anchor-based approaches, such as~\cite{whiteboard_classic2017}, which employs a Faster R-CNN derivative~\cite{faster_rcnn} to detect skewed text in images of whiteboards. On the other, one can make use of pixel-level segmentation, as in~\cite{whiteboard_cnn2018}, which uses SegLink~\cite{seglink}.
Alternatively, for printed text,~\cite{e2e_tables_dengel} introduces robust heuristics on top of the Tesseract OCR engine. Furthermore,~\cite{moysset_whereToStart} relies on a pair of multi-dimensional LSTMs to extract and recognize printed and handwritten text jointly.
Here, as discussed in Section~\ref{sec:method:system}, we follow an anchor-based approach, which 
we found to be effective.